\documentclass[]{fairmeta}
\usepackage{microtype}
\usepackage{amsfonts}
\usepackage{graphicx}
\usepackage{booktabs}
\usepackage{wrapfig}
\usepackage{amsmath}
\usepackage{amsthm}
\usepackage{pifont}
\usepackage{hyperref}
\usepackage[capitalize,noabbrev]{cleveref}
\theoremstyle{plain}

\theoremstyle{definition}

\theoremstyle{remark}

\usepackage{enumitem}
\usepackage[subtle, mathdisplays=tight, charwidths=tight, leading=normal]{savetrees}
\usepackage{booktabs} 
\usepackage{longtable} 
\usepackage{textcomp} 
\usepackage{xspace} 
\usepackage{hyperref}

\newcommand{\distillsparse}{{\normalfont\scshape DistillSparse}\xspace}

\usepackage[table]{xcolor} 

\usepackage{tikz} 

\usepackage{tikz} 
\usepackage{algorithm}
\usepackage{algorithmic}
\usepackage{adjustbox}

\title{Sparrow: \underline{Spar}se \underline{Roll}out for Stable and Efficient Long-context RL of Large Language Models}

\definecolor{skyblue}{RGB}{135, 206, 235} 
\definecolor{palegreen}{RGB}{152, 251, 152}
\definecolor{lightpink}{RGB}{255, 220, 235}
\author[1]{Yang Zhou}
\author[1]{Ranajoy Sadhukhan}
\author[2]{Zhaofeng Sun$^\dagger$}
\author[1]{Zhuoming Chen}
\author[3]{Souvik Kundu}
\author[4]{Saket Dingliwal}
\author[4]{Sai Muralidhar Jayanthi}
\author[4]{Aram Galstyan}
\author[1]{Haizhong Zheng}
\author[1]{Beidi Chen}
\affiliation[1]{Carnegie Mellon University}
\affiliation[2]{Cornell University} 
\affiliation[3]{Intel}
\affiliation[4]{Amazon AGI} 
\contribution[\dagger]{Work done as an intern at Carnegie Mellon University.\\
\texttt{yangzho6@andrew.cmu.edu, rsadhukh@andrew.cmu.edu, zs453@cornell.edu, zhuomingc@andrew.cmu.edu, souvikk.kundu@intel.com, skdin@amazon.com, saimucja@amazon.com, argalsty@amazon.com, haizhonz@andrew.cmu.edu, beidic@andrew.cmu.edu}} 

\abstract{
Despite being powerful, reinforcement learning with verifiable rewards (RLVR) induces extremely long COT, thus making it highly computationally expensive. 
Since RLVR per-step training cost is dominated by long-context generation in rollout, sparse attention offers a promising way to accelerate dense rollout generation. 
However, using sparse rollouts in practice requires a delicate stability-efficiency tradeoff - either sparsity is overly aggressive, leading to collapse, or overly lenient, leading to insufficient speedup. 
In this work, we study the optimal RL stability-efficiency tradeoff through the lens of sparse-to-dense actor-policy mismatch. 
We first observe that \textbf{sparse rollout collapse is not driven by uniform degradation across all tokens}: most tokens generated by sparse rollouts align with dense rollouts perfectly, even under aggressive sparsity. 
Motivated by the observation, we hypothesize that sparse rollout training remains stable as long as the lower tail of the per-token actor-policy mismatch is kept above a critical threshold throughout the sparse rollout trajectory. 
Then, we introduce a \textbf{dynamic sparsity scheduling} technique that maintains this tail statistic constant throughout generation and empirically validate our hypothesis. 
Surprisingly, across a range of model sizes in the Qwen3 thinking family, keeping the tail distribution mismatch statistics at a roughly consistent threshold generally enables stable training. 
We then use a cost-model analysis to find a sparsity schedule for maximum speedup under the mismatch threshold, thus achieving \textbf{2.2x}, \textbf{2.4x}, and \textbf{2.0x} in rollout when training Qwen3-1.7B, Qwen3-4B, and Qwen3-8B. 
Empirically, we show the identified thresholds generalize to a much larger model size (Qwen3-14B) and another RL domain (Coding) and enable stable training. 
Additionally, our analysis naturally motivates the technique \distillsparse. Through lightweight LoRA-based distillation directly on sparse rollout, much more aggressive sparsity can now attain the same sparse-to-dense mismatch threshold, thus achieving higher speedup. 
} 

\metadata[Github]{\url{https://github.com/Infini-AI-Lab/Sparrow}}  
\metadata[Website]{\url{https://infini-ai-lab.github.io/sparrow_project_release/}} 

\begin{document}

\maketitle 

\section{Introduction} 

\begin{figure}[t]
  \centering
  \setlength{\fboxsep}{0pt} 
  \includegraphics[width=\linewidth]{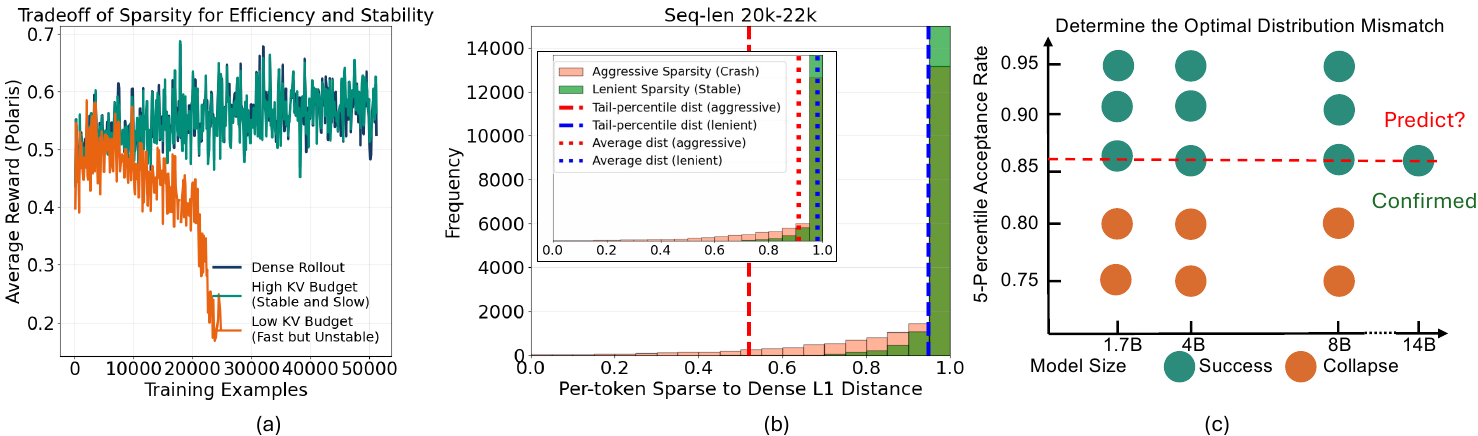} 
  \caption{We study the tradeoff between training stability and efficiency using sparse rollout. (a) illustrates the main problem: arbitrary sparsity tends to miss the optimal tradeoff of lowest cost given stable training. (b) shows a case study demonstrating our insights, in the 20K--22K length bucket, regardless of lenient or aggressive sparsity, the per-token acceptance rate is long-tailed and heavily centered around 1 (top-left figure). The high skewness of the distribution pushes its average acceptance rate toward one, making it less informative to evaluate the distribution. Instead, we choose 5-percentile per-token acceptance rate statistics to focus on the tail distribution as the basis of our study (big figure). Panel (c) displays the main problem: what is the lowest distribution alignment needed to enable stable dense-policy training. For thinking-model sizes 1.7B, 4B, and 8B, we train each model with sparse rollout targeting various levels of distribution mismatch, thus determining the optimal distribution mismatch.} 
  \label{fig:tradeoffandmismatchcost} 
\end{figure}

Reinforcement Learning with Verifiable Rewards (RLVR) has recently driven state-of-the-art performance in mathematical reasoning and code generation \citep{openai2026openaio1card,guo2025deepseek}, but its training cost is extremely high and dominated by long CoT rollout generation, over 70\% of per-step cost \citep{wu2025rlboost}. 
Agentic tasks \citep{anthropic_claude_code_2026,team2025kimi,su2025toolorchestra} further exacerbates the bottleneck, as the rollout now is even longer and beyond max context length. 
\textbf{Sparse attention} can effectively accelerate long-context generation \citep{tang2024quest,sun2024shadowkv,yuan2025native} and thus can replace dense rollout to train the standard dense-attention policies. 
However, naively using sparse rollout for dense policies creates a \textbf{stability-efficiency tradeoff}: aggressive sparsity risks huge policy mismatch and training collapse, while mild sparsity brings RL stability with limited efficiency gains. (Figure~\ref{fig:tradeoffandmismatchcost} (a)) 
This motivates our central question: \textbf{how can sparse rollouts achieve maximal speedup while preserving the stability of dense-rollout RL training across model scales?}

Prior work falls short of addressing this problem. 
(1) \textbf{Sparse Attention Inference Accuracy $\uparrow$ $\neq$ RL Stability $\uparrow$}: many sparse attention techniques have been developed to improve the efficiency--accuracy tradeoff for downstream-task inference \citep{tang2024questqueryawaresparsityefficient,sun2024shadowkv,xiao2024efficientstreaminglanguagemodels,sadhukhan2025kineticsrethinkingtesttimescaling,zhang2023h2oheavyhitteroracleefficient}. 
However, sparse rollout RL instability is mainly due to per-token distribution mismatch between sparse actor and dense policy rather than insufficient rollout rewards. 
(2) \textbf{Suboptimal Convergence Under Severe Actor--Policy Mismatch}: prior work (\cite{liu2025flashrl}, \cite{jackpot2025github}, \cite{luo2026sparse}) addresses actor--policy distribution mismatch, but often studies milder scenarios, such as staleness, where actor--policy KL divergence is one order of magnitude smaller than in sparse rollout. 
When directly applied, these techniques require clipping or masking significant training signals to maintain stability, leading to poor training convergence. 
(3) \textbf{Poor Efficiency}: Sparse rollout stability can be trivially recovered by applying elementwise Top-$k$ or using a huge KV budget, but these approaches do not achieve large efficiency gains. 

Ideally, we desire RL with sparse rollout to (1) enable stable dense-policy training, (2) match dense performance regardless of model size and generation length, and (3) achieve strong efficiency benefits over dense rollout. 

Our study is heavily inspired by the following insight.
The key observation is that \textbf{sparse rollout collapse is not driven by a uniform degradation across all tokens}. Even under aggressive sparsity, most generated tokens remain nearly distribution-aligned with the dense policy. The unstable signal instead appears in the small fraction of tokens where sparse and dense behavior diverge. 
As shown in Figure~\ref{fig:tradeoffandmismatchcost}(b), we collect tokens from sparse-attention generations with 20K prompt length and 2K generation length, then measure each token's distribution mismatch against the dense model. The resulting distribution is highly skewed: most tokens are close to perfectly aligned, even when the sparsity budget is aggressive enough to cause RL collapse. (In fact, in the paper we show that the tail distribution can be well-modeled by a Beta Distribution) \textbf{The high skewness makes average mismatch a weak stability indicator}. For example, when training Qwen3-1.7B with a 37K generation budget, a KV budget of 4096 remains stable while 2560 collapses, but their average per-token sparse-dense L1 distances are still very close and very sensitive to measurement precision: 0.977 and 0.968.
We therefore evaluate sparse-dense mismatch with \textbf{lower-tail statistics} rather than the average. In particular, we use the lower 5-percentile per-token L1 distance to measure the worst-aligned tokens while ignoring the many "perfect" tokens that are unlikely to cause RL training collapse. 
From above formulation, we then hypothesize the following. 
\begin{tcolorbox}[myquote]
\itshape \textbf{Hypothesis}: if the tail distribution mismatch between the sparse actor and dense policy stays above a threshold throughout rollout, sparse rollout for dense policy RL will be stable. 
\end{tcolorbox} 

However, there are \textbf{multiple challenges} in verifying this hypothesis. First, keeping per-token sparse-dense distribution mismatch across the entire trajectory is not straightforward. As shown in Figure~\ref{fig:problem}(b), under any fixed sparsity budget, the per-token mismatch naturally deteriorates as the generation length increases. Second, after holding mismatch statistics constant throughout the trajectory, it remains difficult to know which threshold for the mismatch to target and how it scales as model size increases. Third, even if the above hypothesis holds, as shown in Figure~\ref{fig:tradeoffandmismatchcost}(c), how to achieve the mismatch target with minimal inference cost remains non-trivial. 

\begin{samepage}
In Section \ref{methodcostthreshold}, we systematically study sparse long-context rollout from the perspective of sparse-to-dense distribution mismatch and tame the RL instability when training dense policies to unleash the efficiency benefits from sparse-attention rollout. We use a range of model sizes from Qwen3 thinking family of models under extended RL training with a large and highly challenging math dataset, while enabling rollout length cutoff of 37K\footnote{The maximum model length for Qwen3 family models with 3K problem prompt length.} to approximate real-life RL settings. 
\begin{itemize}[itemsep=0.0pt,topsep=0pt,leftmargin=*]
    \item In Section~\ref{optimalalpha}, we first introduce a technique called \textbf{dynamic sparsity scheduling}, which adjusts the sparsity budget to be more lenient as generation length increases, thus helping us maintain per-token tail mismatch statistics constant throughout rollout of 37K generation length. 
    \item Using this technique, we conduct a systematic controlled study of the relationship between training stability and various trajectory-level tail mismatch statistics. Across a range of models from the same Qwen3 thinking-model family, we search for a threshold of the tail statistics with each model that just keeps RL training stable, lower than which will crash. 
    We validate our hypothesis and find that the optimal threshold is consistent across model sizes.
    \item Then, in Section~\ref{optimalalpha}, under this optimal threshold, we formulate cost models to realistically and reliably compare and analyze dense and sparse rollout. We then use the cost model to derive a practical scheduling guideline for maximizing speedup under the identified per-token mismatch threshold for each model sizes. 
\end{itemize}

Moreover, in Section~\ref{empiricalstudies}, we put our analysis from the control study under extensive empirical validation. 
In Section~\ref{empirical:sparse_rollout_speedup}, across the Qwen3 thinking family, keeping the tail mismatch level above the a consistent threshold enables stable extended-step training, while dynamic sparsity scheduling achieves 2.2$\times$, 2.4$\times$, and 2.0$\times$ generation speedups for Qwen3-1.7B, Qwen3-4B, and Qwen3-8B, respectively. 
In Sections~\ref{generalizel} and~\ref{generalizec}, we apply the same stability criterion to Qwen3-14B thinking-model RL and coding RL, where sparse rollout achieves performance on par with dense rollout. 
Following this analysis, in Section~\ref{distillsparsemethodology}, we introduce \distillsparse to obtain substantial additional speedup on top of dynamic sparsity scheduling by enabling more aggressive sparsity while preserving the same stability criterion. 
\end{samepage} 

\section{Related Works and Problem Statement} 

In this section, we delve into the most relevant related works to our study and cleanly present the problem of interest. 
For a more verbose and comprehensive review of related works, please refer to the extended related work in Appendix \ref{extended_related_works}. 

- \textbf{Prior Rollout Speedup Methods.} 
Despite many recent works proposed to reduce the cost of LLM RL, they have key limitations. 
(1) \textbf{asynchronous RL}~\citep{zheng2025prosperity,piché2025pipelinerlfasteronpolicyreinforcement,zhou2025aprilactivepartialrollouts}  training systems are proposed to accelerate training by decoupling the rollout and training phases to remove bubbles and achieve higher compute utilization. 
However, they cannot reduce the underlying long-context generation workload; when the rollout workload is large, training can remain slow even at high utilization. 
(2) while \textbf{model quantization in RL}~\citep{liu2025flashrl,huang2025qerlefficiencyquantizationenhanced} is proposed to reduce the cost of loading model weights, it cannot effectively mitigate the rollout overhead for long-sequence generation, where KV-cache loading remains the primary bottleneck~\citep{sadhukhan2025kineticsrethinkingtesttimescaling}.
(3) Conversely, \textbf{speculative decoding} (SpecDec)~\citep{leviathan2023fastinferencetransformersspeculative, chen2023acceleratinglargelanguagemodel, liu2026specrlacceleratingonpolicyreinforcement, he2025historyrhymesacceleratingllm} can accelerate rollouts without altering the sampling distribution. Still, they suffer from weaknesses. First, the SpecDec techniques requiring draft models cannot achieve substantial practical speedup under large batch-size settings, as verification becomes compute-intensive per \citep{liu2025turbospecclosedloopspeculationcontrol,su2023synergyspeculativedecodingbatching}. 
Second, the SpecDec techniques using past epochs' rollout generation as a draft to speed up new epoch rollout are not always practically applicable, as realistic industry RL training is usually done in one epoch. 

\begin{figure}[t]
  \centering
  \setlength{\fboxsep}{0pt} 
  \includegraphics[width=\linewidth]{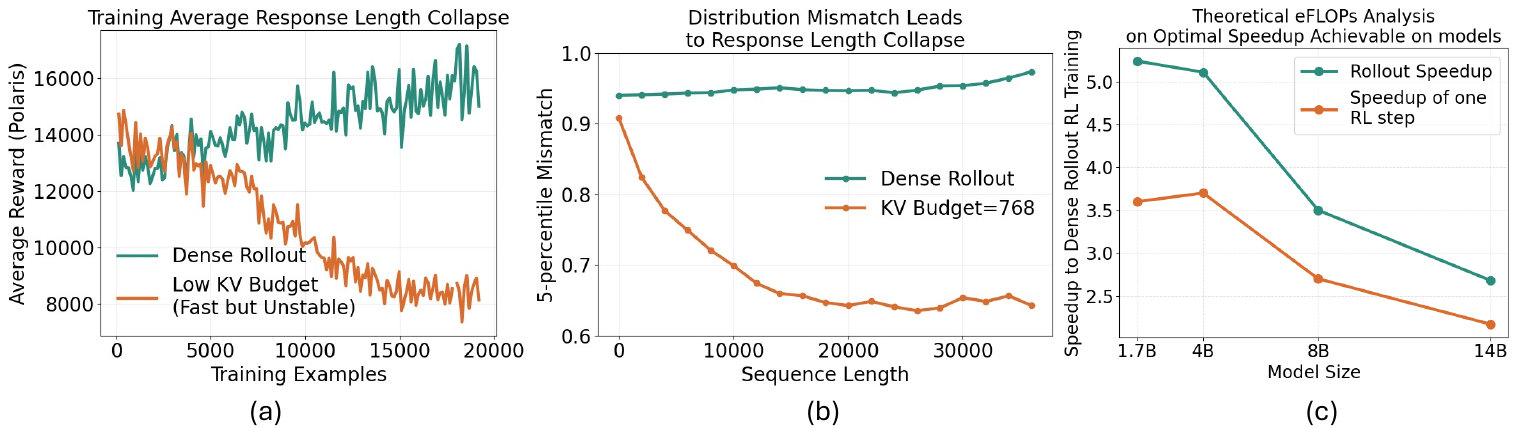} 
  \caption{(a) Aggressive sparsity causes rollout response length to shrink during training, which correlates with dense-training collapse. (b) Unlike dense rollout, aggressive sparsity leads actor--policy distribution mismatch to grow sharply with sequence length. (c) Given that we identified the optimal tail mismatch between sparse rollout and dense training as above 0.86 and under 0.90, we then estimate the theoretical optimal speedup ratio achievable from sparse rollout using our proposed cost model. Cost in (c) is estimated with eFLOPs on H200. We find that as the model size increases, the optimal speedup decreases. The two main reasons are first that the attention contribution to the total generation cost decreases as model size increases, and second that as model size increases, a higher KV budget is required to maintain the actor-policy mismatch above the identified threshold; more details are provided in Section~\ref{costalpha}.} 
  \label{fig:problem} 
\end{figure} 

- \textbf{Sparse attention} is a promising solution to speed up RL rollout. First, during RL training in practice for thinking models, long-context rollout accounts for over 70\% of per-step cost \citep{wu2025rlboost}. 
Second, sparse attention has been extensively studied and proven to be effective in achieving significant speedup with strong preservation of generation quality. \citep{xiao2024efficientstreaminglanguagemodels,zhang2023h2oheavyhitteroracleefficient,deepseekai2024deepseekv32,chen2024magicpig}
For our study, we zoom in on one particular sparse attention family, \textbf{block-sparse attention}, as a case study \citep{lu2025mobamixtureblockattention,yuan2025nativesparseattentionhardwarealigned,tang2024quest,sun2024triforcelosslessaccelerationlong}.\footnote{We believe similar conclusions about sparse rollout dense training can be drawn in fine-grained sparse-attention works.} 
Arbitrarily configuring sparsity in rollout is highly prone to training collapse, and as shown in Figure \ref{fig:problem}(a) the response length tends to tank during collapse.\footnote{For the experiment, we train Qwen3-1.7B thinking model with 37K generation length on Polaris with KV budget 1024.} 
However, prior works in sparse attention over-emphasize generation accuracy, which is very different from the true cause of RL collapse, per-token actor-policy distribution mismatch. 
We showcase the above by running the following experiment. 
In Appendix~\ref{spdnhighreward} we show that under sparse rollout that triggers dense policy collapse, even if we apply test-time-scaling and filtering to drastically increase the average rewards, even above those of dense rollout, the training performance is still not recovered. 

- \textbf{Algorithmic Techniques on Actor-Policy Alignment} have also been extensively studied, but they cannot provide a general solution for maintaining sparse rollout dense training RL stability. 
These techniques are either importance-sampling based \citep{espeholt2018impalascalabledistributeddeeprl,fu2025areal,liu2025flashrl} or rejection-sampling based \citep{jackpot2025github,qi2025defeatingtraininginferencemismatchfp16,luo2026sparse}. 
However, these techniques prioritize conventional actor-policy mismatch scenarios such as staleness. 
However, we show in Appendix~\ref{app:rollout_cost_sparse} that sparse rollout introduces a failure mode that induces orders of magnitude larger KL divergence between actor and policy than staleness problems, where these techniques succeed. 
Thus, these methods fail to stabilize the training shown in Figure \ref{fig:problem}(a).\footnote{We use TIS to correct Qwen3-1.7B thinking RL with sparse rollout with KV budget of 1024 and 37K generation length cutoff. In Appendix \ref{visualizationmismatch}, we also show that rejection-sampling based method also cannot solve the instability under the same condition.} 
Admittedly, we can drastically increase the truncation/rejection threshold for more stability, but this will hurt convergence. 
\textbf{Therefore, to realistically unlock the efficiency benefits of sparse attention in long-context rollout, we need to systematically study the lowest sparse-rollout cost that still enables stable dense-model training, as outlined in the next sections.} 
 
\section{Observation and Insights} 
\label{methodcostthreshold} 

In this section, we present the insights and lay out the detailed study of the relationship between sparse attention and actor-policy distribution mismatch. 
Specifically, we show in \ref{analysisalpha} that even under aggressive sparsity, the majority of tokens exhibit perfect alignment with the dense model, while only a minor portion of tokens behave poorly. 
Based on the observation, in \ref{optimalalpha} we choose per-token metrics of distribution mismatch focusing on the tail and empirically show that the threshold for training stability is consistent across model sizes. 
Furthermore, \ref{costalpha} factors in cost considerations: the minimum sparse-attention cost needed to enable stable dense training. 

\subsection{Analysis on per-token Acceptance Rate: High Skewness Makes Average Acceptance Rate less Meaningful} 
\label{analysisalpha} 
Acceptance Rate $\alpha$ or L1 distance between two distributions is widely used \citep{leviathan2023fastinferencetransformersspeculative} to measure the divergence between two distributions $P$ and $Q$, which can be formulated as follows. 

\vspace{0.1em} 
\begingroup 
\begin{equation}
    \alpha = \mathbb{E}_{x \sim Q}\!\left[\min\!\left(\frac{p(x)}{q(x)}, 1\right)\right] = \mathbb{E}_{x \in V}\!\left[\min(p(x), q(x))\right]
\end{equation} 
\endgroup 
\vspace{0.1em} 

\begin{wrapfigure}{r}{0.35\textwidth}
  \centering
  \setlength{\fboxsep}{0pt}
  \includegraphics[width=0.35\textwidth]{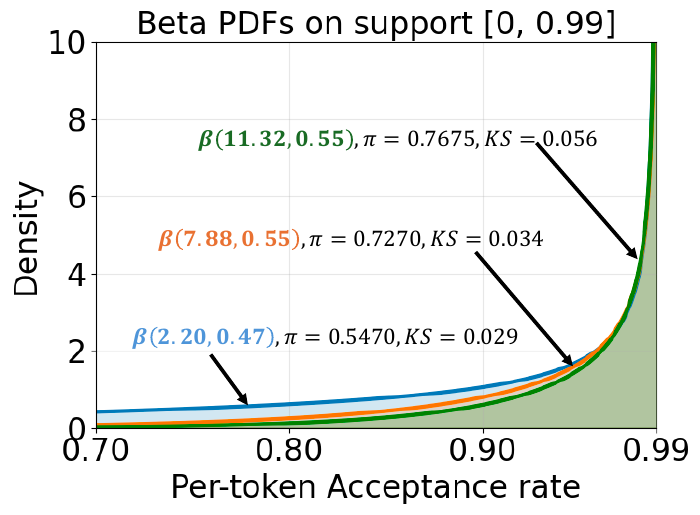}
  \caption{The tail distribution of per-token mismatch between sparse and dense can generally be approximated very well by a Beta distribution. Qwen3-1.7B thinking with three sparsity levels (20K-22K). Budgets are 512 (blue), 8192 (coral), and 16384 (green).} 
  \label{fig:figurein} 
\end{wrapfigure} 

Naturally, given a sparsity configuration of constant KV budget, the acceptance rate decreases as the generation sequence length increases. 
Similarly, when looking at acceptance rate with a specific sequence length, the acceptance rate decreases as the sparsity configuration becomes more aggressive, or the KV budget becomes smaller. 
However, in this section, \textbf{we reveal that simply averaging the per-token acceptance rate given a sparsity configuration and sequence length is not ideal as an estimate of the distribution mismatch between the sparse and dense probability distributions}. 

The core problem is that given a population of tokens collected under the same sparsity configuration and sequence length, the dominant majority of tokens see their per-token acceptance rate above 0.999, perfectly aligned with the dense policy. 
The high skewness towards 1 causes a severe problem: the per-token acceptance rate is almost not correlated with sequence length increases, and the average acceptance rate falls into similar values regardless of aggressive and lenient sparsity. 
We use the Qwen3-1.7B Thinking model with an aggressive KV budget (512) to generate sequences up to length 40K\footnote{We ask the inference engine to also return per-token position Top 20 tokens and logprobs, which enables us to approximate the per-token acceptance rate at every token position.} and compute each token's per-token acceptance rate. 
We calculate the Pearson's r statistic between per-token acceptance rate and sequence length, and the value is consistently around -0.1, suggesting a very weak correlation. 
Similarly, shown in Figure \ref{fig:tradeoffandmismatchcost}, we use KV budgets of 512, 4096, and 16384 and look at all the tokens between 20k and 22k. Even though the KV budget differs significantly, the average acceptance rates in the 20k--22k sequence-length bucket are all pushed close to 1. 
\textbf{In practice, KV budgets of 2560 and 4096 give average acceptance rates of 0.968 and 0.977, but one causes a crash and one does not.} 

Unlike the average acceptance rate, which is less informative when the distribution is highly skewed, statistics that focus on the tail distribution matter much more for evaluating the distribution mismatch between the approximate distribution and the original distribution. 
Notice that besides sharing a similar average, the smaller KV budget distribution possesses a much thicker tail than the higher KV budget. 
Specifically, as shown in Figure~\ref{fig:figurein}, we find that if we leave out all the tokens with per-token acceptance rate higher than 0.999, the tail can be modeled very nicely by a scaled Beta distribution of support [0, 0.999], with all three distributions having a Kolmogorov-Smirnov (KS) statistic under 0.04. 
From the distribution for the run with KV budget 512, the variance of the respective beta distribution is 0.19, much more spread out than 0.05 for KV budget 16384. 

Conceptually, tokens perfectly aligned with dense policy are not the culprit of training collapse.
The issue of training collapse comes from the minority of tokens that are drastically different from dense tokens in probability distribution. 
Therefore, instead of focusing on the average per-token acceptance rate, we focus on statistics that focus on the tail distribution. 
\begin{tcolorbox}[myquote]
\itshape
\textbf{Method}: we choose the 5-percentile cutoff of the per-token acceptance rate for our study to focus on the tail distribution of actor and policy mismatch in RL training. 
\end{tcolorbox} 
Also, the 5-percentile cutoff achieves -0.55 Pearson's r, strongly correlated with sequence length. 
Shown in Figure \ref{fig:proberties}(a), the 5-percentile per-token acceptance rate allows the same sparsity configurations to show consistent mismatch values with the dense policy regardless of training iterations under the baseline dense rollout scenarios, almost the same between iteration 1 and iteration 550. Even though the dense policy improves as training continues, under the same sparsity budget, the mismatch is only relative to the corresponding dense policy. 

\begin{figure}[t]
  \centering
  \setlength{\fboxsep}{0pt}
  \includegraphics[width=\linewidth]{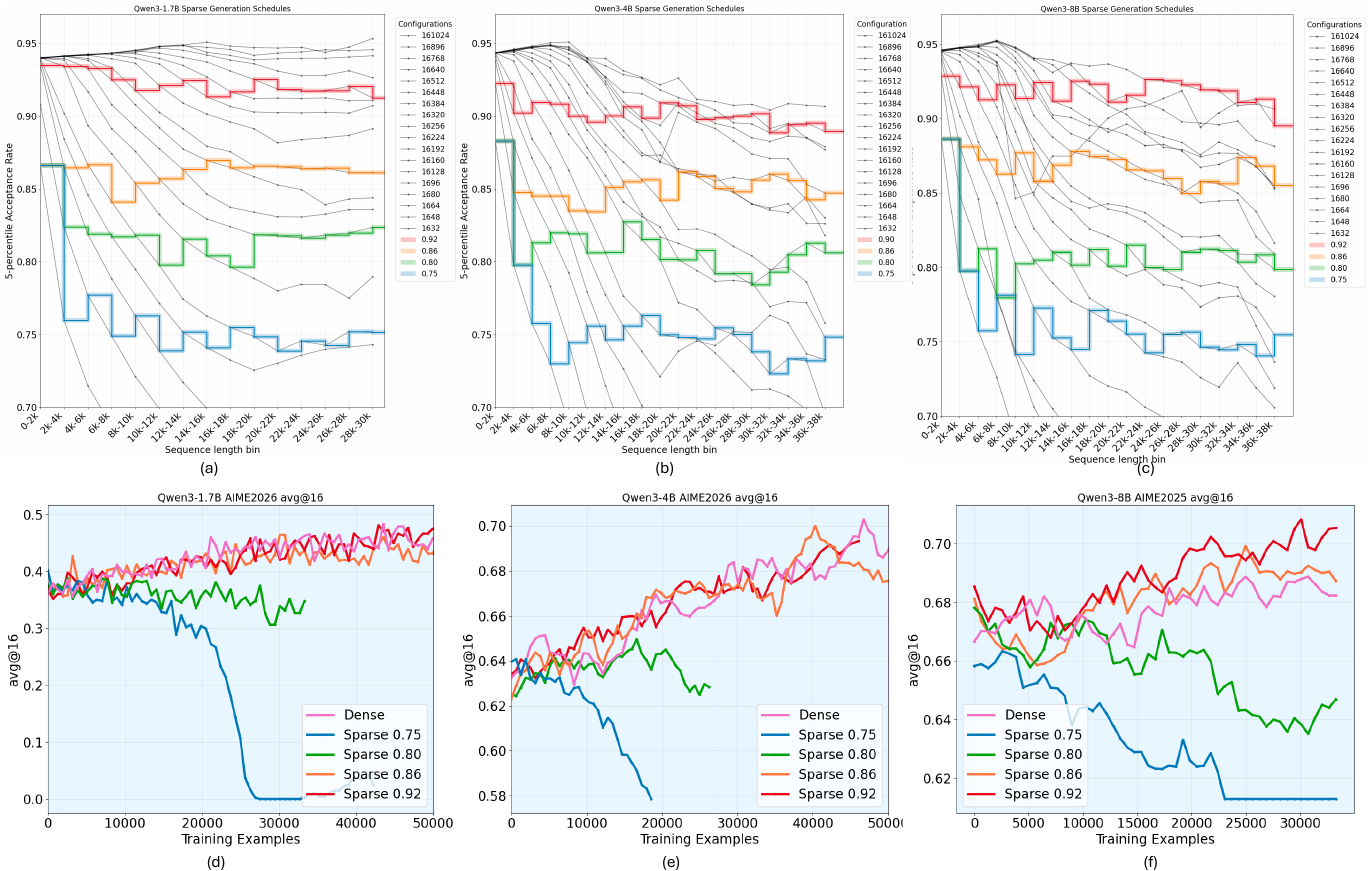} 
\caption{Control Study 1. Each column corresponds to one model: Qwen-1.7B, Qwen-4B, or Qwen-8B. The first row plots the 5-percentile acceptance rate versus sequence length for different sparsity configurations, all of which exhibit decaying acceptance rates as sequence length increases. We determine sparsity schedules that control per-trajectory divergence at 0.75, 0.8, 0.86, and 0.92. With these schedules, the second row evaluates whether each divergence target enables stable training with performance on par with dense rollout.} 
  \label{fig:control-study-grid1} 
\end{figure} 

\subsection{Control Study 1: What Threshold of Actor--Policy Mismatch Enables Stable RL Training, and How Does It Scale with Model Size?} 
\label{optimalalpha} 

The main challenge in cleanly studying the actor-policy mismatch to RL training stability\footnote{For all the RL training runs presented, we use TIS by default. It turns out that 37K generation is so long that even dense rollout dense training will hit the inference decode and training prefill kernel mismatch issue and crash, if it does not use TIS.} with sparse attention is that the mismatch naturally increases as sequence length increases when the sparsity budget is fixed before generation. 
We address it by introducing sparsity scheduling. 
\begin{tcolorbox}[myquote]
\itshape 
\textbf{Method}: We use \textbf{Sparsity Scheduling}, which increases the sparsity budget as the sequence length increases, thus preserving the tail distribution mismatch of the entire generation trajectory bin at a targeted value. 
\end{tcolorbox} 

As shown in Figure \ref{fig:control-study-grid1}(a--c), we first measure the 5-percentile acceptance rate of 18 sparsity configurations for each model across 20 sequence-length bins. All configurations use page size 16 with varying numbers of pages. We consider four distribution-divergence targets: 0.75, 0.80, 0.86, and 0.92. To attain each target, we search horizontally across sparsity settings to find the desired configuration for each divergence target and sequence-length regime. 
Across all three model sizes, we follow similar procedures to find the configurations. 

Specifically, to make sure that our measurement of the per-token acceptance rate statistics is generalizable to math training datasets in general, we use AIME 2024, 2025, and 2026 together, repeating them 8 times as the prompt set and using sparse attention models for generation. The sparse generation with log-prob distributions is computed with dense log-probs for the per-token acceptance rate at each trajectory position. 

Also, as the per-token acceptance rate generally decreases as sequence length increases, to facilitate analysis, we group tokens based on their position sequence length, with 2K as the bin size. 37K generation length encompasses tokens separated into 19 bins. We compute the per-bin tail statistics for our study. 

Then, following the four divergence levels, we ensure that the trajectories sampled by sparse rollout reach a steady 5-percentile acceptance rate. 
We train dense policies using rollouts of four targets to see whether they enable stable dense policy RL, as shown in Figure \ref{fig:control-study-grid1} (d), (e), (f). 
\begin{tcolorbox}[myquote,colback=blue!8,colframe=blue!60!black]
\itshape 
\textbf{Takeaway 1}: Across the Qwen3 thinking model family, the optimal distribution mismatch threshold ($\tau$) is roughly 0.86 and less than 0.92, and $\tau$ is consistent across increasing model sizes. 
\end{tcolorbox} 
The identified threshold suggests stable RL training leaves some room for efficiency improvement, and we will see how to maximize the efficiency benefits next. 

\subsection{Control Study 2: What Is the Lowest Cost Needed for Stable and Performant Dense-Model Training?} 
\label{costalpha} 
To make sure the cost analysis can transfer to different hardware easily, we follow \citep{sadhukhan2025kineticsrethinkingtesttimescaling} and model the cost of rollout based on model size and hardware memory bandwidth. 
For dense attention, we can formulate the compute, memory, and combined cost of repeated sampling N times in eFLOPs (equivalent FLOPS), given model parameters \textit{P}, input prompt length in tokens \textit{$L_{in}$}, output prompt length in tokens \textit{$L_{out}$}, dimension of both Key and Value \textit{D}, GQA ratio (greater than 1) in \textit{r}, and 1 over the GPU SRAM memory bandwidth \textit{I}
\[
C_{comp} = 2 \cdot P \cdot N \cdot L_{out} + r \cdot (2 \cdot L_{in} + L_{out}) \cdot L_{out} \cdot N \cdot D 
\]
\[
C_{mem} = 2 \cdot L_{in} \cdot L_{out} \cdot D + N \cdot L_{out}^2 \cdot D
\] 
\begingroup 
\begin{equation}
C_{dense} = C_{comp} + I \cdot C_{mem}
\label{eq:dense-cost}
\end{equation}
\endgroup 
Similarly, we can model the same for block-sparse attention as follows. For simplicity, we assume that Top-$k$ kernels incur only minimal overhead for \texttt{pagesize} $\geq$ 16. Sparsity uses KV budget \textit{B} and \textit{pagesize}, 
\[
C_{\text{sparse, no scoring}} =
\underbrace{2 \cdot N \cdot P \cdot L_{out} + 2 \cdot r \cdot N \cdot D \cdot B \cdot L_{out}}_{\text{compute}}
+
\underbrace{2 \cdot I \cdot N \cdot D \cdot B \cdot L_{out}}_{\text{memory}} .
\] 
\[
C_{scoring} =
\underbrace{2 \cdot N \cdot L_{in} \cdot D \cdot L_{out} + \frac{r \cdot N \cdot D \cdot L_{out}^{2}}{2 \cdot \text{pagesize}}}_{\text{compute}}
+
\underbrace{2 \cdot I \cdot L_{in} \cdot D \cdot L_{out} + \frac{I \cdot N \cdot D \cdot L_{out}^{2}}{2 \cdot \text{pagesize}}}_{\text{memory}} .
\] 
\begingroup 
\begin{equation}
C_{sparse} = C_{\text{sparse, no scoring}} + C_{scoring}
\label{eq:sparse-cost}
\end{equation}
\endgroup 


\begin{figure}[t!]
  \centering
  \includegraphics[width=\textwidth]{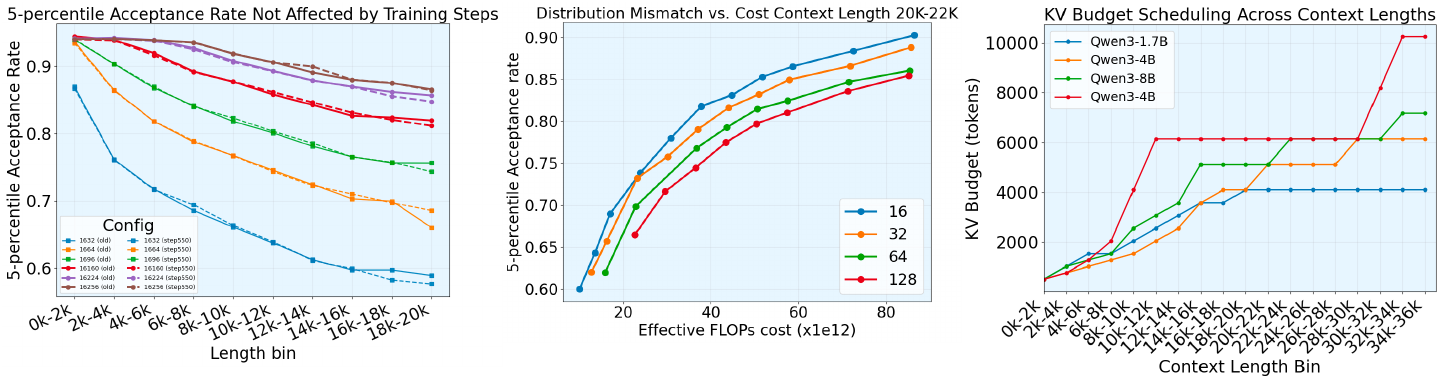} 
  \caption{(a) shows an important property of the 5-percentile acceptance rate: it is not dependent on training iteration, and is instead determined by the dense model, sparsity level, and sequence length. (b) shows that page size 16 leads larger page sizes in the tradeoff between distribution alignment and efficiency. (c) shows that larger models generally favor larger KV budgets to maintain the same distribution-alignment level (0.86 here).} 
  \label{fig:proberties} 
  
\end{figure} 

Following the above cost-model formulation, we search over page sizes 16, 32, 64, and 128 for each model size: 1.7B, 4B, and 8B.\footnote{page size 8 or below requires much more system optimization and is currently not supported by our sparse inference engine. Future work will extend this analysis to smaller page sizes.} 
Based on the proposed cost model, we identify an important property of block-sparse attention: for page sizes greater than or equal to 16, page size 16 dominates the Pareto frontier for the tradeoff between tail acceptance rate and cost. Figure \ref{fig:proberties}(b) shows this scenario for the 1.7B model at sequence lengths 20k--22k. 
The pattern holds true across other length regimes as well; more details are provided in Appendix \ref{costbreakdown}. 
The core reason behind this phenomenon is that from page size 16 onwards, the cost of computing landmark scores and picking Top-$k$ is not expensive and much less significant than the actual attention computation, as predicted by the cost model. 

\begin{tcolorbox}[myquote,colback=blue!8,colframe=blue!60!black]
\itshape 
\textbf{Takeaway 2}: For block-sparse attention (\texttt{pagesize} $\geq$ 16), more fine-grained sparsity (\texttt{pagesize} $\downarrow$) consistently beats the coarser sparsity (\texttt{pagesize} $\uparrow$) in distribution mismatch with the dense model across all cost regimes. 
\end{tcolorbox} 

Importantly, beyond the pattern predicted on paper by the cost model, we show in Appendix \ref{costbreakdown} Table \ref{tablecostpage} that if the Top-$k$ kernels are implemented properly, the empirical cost of sparse attention model generation is insensitive to page-size selection. 
Therefore, we know that to achieve the target acceptance rate with the smallest cost, we stick to using the uniform page size of 16 for sparsity going forward. 
 
\section{Empirical Studies} 
\label{empiricalstudies} 

\begin{table*}[t]
\centering 
\caption{
Comprehensive Evaluation of Sparse Rollout relative to Dense, where we report the best-achieved scores under each downstream task and task-related metrics. 
}
\label{tab:bigbigeval} 
\small
\setlength{\tabcolsep}{4pt}
\begin{adjustbox}{max width=\linewidth}
\begin{tabular}{lcccccccc}
\toprule
\textbf{Models} 
& \textbf{AIME25} 
& \textbf{AIME25} 
& \textbf{AIME26} 
& \textbf{AIME26} 
& \textbf{AIME24} 
& \textbf{AIME24} 
& \textbf{AMC22/23} 
& \textbf{AMC12} \\
& \textbf{Mean@16} 
& \textbf{Pass@16} 
& \textbf{Mean@16} 
& \textbf{Pass@16} 
& \textbf{Mean@16} 
& \textbf{Pass@16} 
& \textbf{Mean@4} 
& \textbf{Mean@4} \\
\midrule

\multicolumn{9}{c}{
\textit{Qwen3-1.7B Thinking} 37K generation length cutoff; one epoch on 53K Polaris examples; 
} \\
\midrule
Dense Rollout Dense Training & 0.4322 & 0.5440 & 0.4833 & 0.5995 & 0.5687 & 0.7172 & 0.8072 & 0.7333 \\
Sparsity setup (0.92) Rollout & 0.4625 & 0.5729 & 0.4875 & 0.6197 &  0.5770 & 0.7155 & 0.834 & 0.7222 \\
\rowcolor{cyan!10} 
Sparsity setup (0.86) Rollout & 0.4145 & 0.5436 & 0.4645 & 0.5914 & 0.5416 & 0.6958 & 0.8192 & 0.7166 \\
\rowcolor{cyan!10} 
\textbf{Sparse with LoRA Distillation Sec. \ref{distillsparsemethodology}} & 0.4354 & 0.5562 & 0.4562 & 0.5874 & 0.5333 & 0.6929 & 0.7333 & 0.8072 \\
\midrule
\multicolumn{9}{c}{
\textit{Qwen3-4B Thinking} 37K generation length cutoff; one epoch on 53K Polaris examples; 
} \\
\midrule
Dense Rollout Dense Training & 0.7145 & 0.7933 & 0.7229 & 0.8216 & 0.775 & 0.836 & 0.8 & 0.9367 \\
Sparsity setup (0.92) Rollout & 0.7 & 0.7978 & 0.7062 & 0.8158 & 0.7729 & 0.836 & 0.7944 & 0.9337 \\
\rowcolor{cyan!10} 
Sparsity setup (0.86) Rollout & 0.7416 & 0.8321 & 0.7125 & 0.8197 & 0.756 & 0.832 & 0.7944 & 0.9397 \\
\midrule
\multicolumn{9}{c}{
\textit{Qwen3-8B Thinking} 37K generation length cutoff; one epoch on 33K Polaris examples; 
} \\
\midrule 
Dense Rollout Dense Training & 0.7083 & 0.8000 & 0.6979 & 0.8176 & 0.7770 & 0.8637 & 0.8111 & 0.9337 \\
Sparsity setup (0.92) Rollout & 0.7229 & 0.8076 & 0.725 & 0.8233 & 0.7815 & 0.8456 & 0.8111 & 0.927 \\
\rowcolor{cyan!10} 
Sparsity setup (0.86) Rollout & 0.7145 & 0.8055 & 0.6979 & 0.8125 & 0.7791 & 0.8441 & 0.8277 & 0.9216 \\ 
\midrule 
\multicolumn{9}{c}{
\textit{Qwen3-14B Thinking} 37K generation length cutoff; one epoch on 53K Polaris examples; 
} \\
\midrule 
Dense Rollout Dense Training & 0.75 & 0.8269 & 0.7770 & 0.8646 & 0.8125 & 0.8810 & 0.8055 & 0.9396 \\
\rowcolor{cyan!10} 
Sparsity setup (0.86) Rollout & 0.75 & 0.828 & 0.7708 & 0.8558 & 0.8028 & 0.8696 & 0.8111 & 0.9367 \\
\midrule 

\end{tabular} 
\end{adjustbox} 
\end{table*} 

In this section, we evaluate sparse rollout across four empirical axes. 
\ref{empirical:sparse_rollout_speedup} shows that dynamic sparsity scheduling maintains dense-level RL performance while achieving practical rollout speedups. 
\ref{generalizel} tests generalization: whether the same threshold transfers to Qwen3-14B too large and infeasible for exhaustive sparsity sweeps. 
Similarly, \ref{generalizec} show our hypothesis generalizes to coding RL. 
In summary, we are able to achieve 2.2x, 2.4x, 2.0x, 1.48x rollout speedup for Qwen3-1.7B, Qwen3-4B, Qwen3-8B, and Qwen3-14B models, while seeing our method generalize to coding RL. 
Finally, Section~\ref{distillsparsemethodology} presents \distillsparse, which is naturally motivated by our analysis, further enabling more aggressive sparse rollout with significantly more speedup. 

\subsection{Sparse Rollout Maintains RL Training Stability while Achieving Practical Speedup} 
\label{empirical:sparse_rollout_speedup} 

\begin{wraptable}{r}{0.6\textwidth} 
\scriptsize 
\caption{Throughput and speedup from dynamic sparsity scheduling, benchmarked on 4\texttimes H200 GPUs.} 
\label{tab:dynamic_sparsity_speedup} 
\setlength{\tabcolsep}{5pt}
\renewcommand{\arraystretch}{1.08}
\begin{tabular}{@{}lcccc@{}}
\toprule
\textbf{Model} & \textbf{Dense t.$\uparrow$} & \textbf{Sparse t. $\uparrow$} & \textbf{Rollout} & \textbf{RL One-step} \\
& \textbf{(tokens/s)} & \textbf{(tokens/s)}  & \textbf{Speedup} & \textbf{Speedup} \\
\midrule
Qwen3-1.7B (0.86) & 2561 & 5662 & 2.21$\times$ & 1.97$\times$ \\
\midrule
Qwen3-4B (0.86) & 1369 & 3308 & 2.41$\times$ & 2.11$\times$ \\
\midrule
Qwen3-8B (0.86) & 1300 & 2682 & 2.06$\times$ & 1.81$\times$ \\
\midrule 
Qwen3-14B (0.86) & 1544 & 2289 & 1.48$\times$ & 1.35$\times$ \\
\bottomrule 

\end{tabular} 
\end{wraptable} 

Following the plan laid out in \ref{optimalalpha}, we train Qwen3 \citep{yang2025qwen3technicalreport} family models 1.7B, 4B, 8B for one epoch on Polaris \citep{Polaris2025} using the planned sparsity schedule based on the tail acceptance rate statistics (5-percentile acceptance rate). 
Specifically, to enable high infrastructure utilization during training, we implement the dynamic sparsity scheduling generation required by our design based on open-sourced sparse attention library Vortex \citep{vortex_torch2026}. 

The results are summarized in Table \ref{tab:bigbigeval} with tests of three different years 2024, 2025, 2026 of AIME problems, each averaged over 16 times, and two years 2024, 2023 of AMC problems each averaged over 4 times, showing that they are on par with dense. 

\textbf{Implementation Efficiency} - To mimic the distribution of rollout during RL training of Polaris, we gather AIME24 and repeat every problem 8 times with 32K generation length cutoff. 
Results are shown in Table \ref{tab:dynamic_sparsity_speedup}.
For profiling, we deploy 4xH200 GPUs. 
We show that across 1.7B, 4B, and 8B, we can achieve over 2.0x speedup on RL rollout, which is around 2.0x speedup on the end-to-end iteration time. 

\subsection{Testing Threshold Generalization on Much Larger Model} 
\label{generalizel} 

\begin{figure}[t] 
\includegraphics[width=\linewidth]{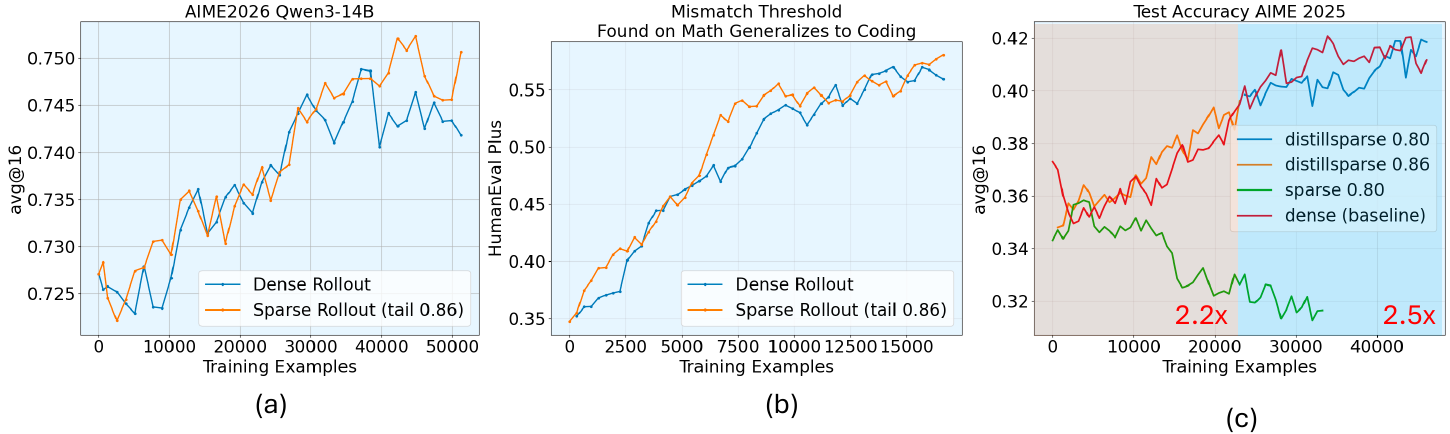} 
\caption{(a) We show the training stability of the Qwen3-14B Thinking model. Under a 37K generation length cutoff on Polaris, rollout with the 0.86 sparsity setting achieves downstream results on par with dense rollout. (b) Sparse rollout with the identified threshold also enables stable RL training beyond math data, matching dense-rollout performance on coding. (c) \distillsparse provides an opportunity to maintain training stability while gaining additional speedup.} 
\label{mut} 
\vspace{-0.2cm}
\end{figure} 

\begin{wraptable}{r}{0.55\textwidth}
\centering
\scriptsize
\caption{Coding RL evaluation for Qwen3-1.7B Thinking with a 24K generation length cutoff after one epoch on 21K filtered TACO.} 
\label{tab:coding_generalization} 
\setlength{\tabcolsep}{3pt}
\renewcommand{\arraystretch}{1.08}
\begin{adjustbox}{max width=\linewidth}
\begin{tabular}{lccc}
\toprule
\textbf{Setups}
& \textbf{LiveCodeBench}
& \textbf{HumanEval+}
& \textbf{MBPP+} \\
\midrule
Dense Rollout Baseline & 0.3970 & 0.6219 & 0.5846 \\
\rowcolor{cyan!10} 
Sparsity (0.86) & 0.3933 & 0.6280 & 0.6084 \\
\bottomrule
\end{tabular}
\end{adjustbox}
\end{wraptable} 
Beyond the three model sizes for which we have enough compute to sweep sparsity settings, we explore ways to extend our study to regimes where enumeration is no longer feasible. 
A Qwen3-14B Thinking model with a full 37K generation cutoff takes 4 H200 nodes (32 H200s) and 8 full days (~190 hrs) to train for an entire epoch. For us, it is no longer possible to brute-force all potential options for the optimal sparsity schedule that maximize the efficiency benefit while preserving training convergence. 

We follow the 5-percentile acceptance rate guidance to keep 0.86 and train Qwen3-14B using the determined sparsity schedule. The training runs for the epoch on Polaris. 
The dense policy from sparse rollout matches the dense policy from dense rollout RL very closely. 
We show the average training reward figure in Figure \ref{mut}(a), and we show that the average training reward is highly close to dense reward with gap consistently less than 1\%. We also populate the downstream evals in Table \ref{tab:bigbigeval}, showing that the Qwen3-14B matches the dense rollout training convergence. 


\subsection{Testing Threshold Generalization on Coding RL} 
\label{generalizec} 

Beyond math reasoning, training strong coding model also requires RL to scale up CoT to achieve higher pass-rate in harder coding problems. 
We show that the threshold found using math domain can be directly used for coding training. 
For the experimental implementation, we follow \cite{code-r1} to set up a sandbox for reward computation and use the 21K filtered TACO training examples from \cite{li2023taco} to train a Qwen3-1.7B Thinking model with the 0.86 sparsity setup. 
We evaluate the training performance with LiveCodeBench \citep{han2024livecodebench}, HumanEval+, and MBPP+ \citep{evalplus}. 

The results are reported in Table \ref{tab:coding_generalization}, where we show that the threshold determined in math reasoning generalizes to other generation workloads, coding. 
We also present RL training convergence in Figure \ref{mut}(b), showing that sparse rollout matches dense-rollout convergence over extended training. 

\subsection{Optimal KV Budget Scales up as Model Sizes Increases} 
\label{empirical:kv_budget_scaling} 
An important phenomenon we discovered is that as model size increases, the KV budget needed for generation at any sequence length also increases in order to hold the tail acceptance rate constant. 
In Figure ~\ref{fig:proberties}, we show that across sequence length, 14B (red) requires the most KV budget to attain 5-percentile acceptance rate 0.86, followed by 8B (green), 4B (yellow), and 1.7B (blue). 
This result also contributes to the decreasing trend of optimal speedup achievable by these models shown in Figure \ref{fig:tradeoffandmismatchcost}(c) and explains the speedup shown for 14B in Table \ref{tab:dynamic_sparsity_speedup} using sparse attention.
 

\subsection{Case Study \distillsparse: Achieving the Same Stability Threshold with Lower Sparse-Rollout Cost} 
\label{distillsparsemethodology} 

\begin{wrapfigure}{r}{0.5\textwidth}
    \centering
    \vspace{-0.5em}
    \includegraphics[width=\linewidth]{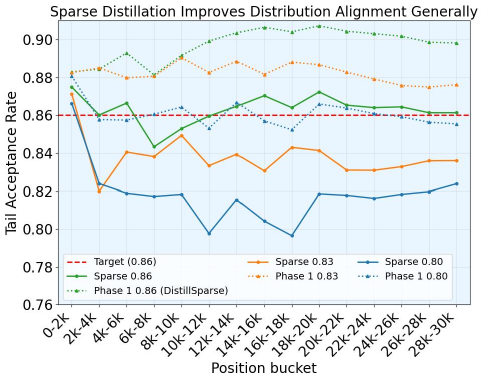} 
    \caption{\distillsparse training on sparsity 0.86 after 20K training examples. We show that \distillsparse generally improves sparse attention toward the dense distribution, as shown by the distribution alignment of sparsity 0.80 and 0.83. In fact, the old sparsity setting with 0.80 tail mismatch now gets boosted to 0.86, close to the identified threshold.} 
    \label{fig:original_vs_distillsparse_step90}
    \vspace{-1em}
\end{wrapfigure} 

Although threshold studies enable us to understand the actor-policy divergence and acquire the best tradeoff between stable training and optimal efficiency, it is still important to explore opportunities to go beyond the optimal efficiency from the threshold for a higher ceiling of efficiency improvement. 
Among potential techniques, we argue that one straightforward opportunity is to actively bring the sparse rollout closer to the dense model through lightweight distillation. 
Sparse rollout already generates extended trajectories, while dense policy training already requires computation of dense log-probabilities. 
Without significant additional cost, we can reuse the computation to actively fine-tune the sparse rollout directly with on-policy distillation~\citep{agarwal2024policy}. 

Then, we advocate for a LoRA-based distillation called \distillsparse that goes beyond the previous limit. 
The complete description of each step of RL training is described in Algorithm \ref{algo:lora-distill}. 
The on-policy distillation process is naturally viable: no extra computation is required for new sparse student trajectory generation, and the dense attention teacher log-probs used to compute dense policy update (step 4 in Algorithm \ref{algo:lora-distill}) can be repurposed to compute the distillation loss in step 7. 
The core challenge is to distill the sparse rollout without contaminating the training signal of the dense policy weights.
Thus, we propose training a set of LoRA weights for aligning sparse and dense models that are added only during rollout (sparse generation), discarded during dense policy log-probability computation and dense policy training. 
\textbf{Next, we illustrate that with distillation of \distillsparse, sparse attention is more aligned with dense attention in general and enables much more aggressive sparsity to attain the same distribution mismatch threshold we found to enable stable RL training with higher speedup.} 


\begin{algorithm}[t]
\caption{Online LoRA distillation for sparse RL training}
\label{algo:lora-distill}
\begin{algorithmic}[1]
\REQUIRE Base model parameters $\theta$; LoRA parameters $\phi$; prompt batch $x \sim \mathcal{D}$
\STATE Construct \textbf{dense policy} $\pi^{\text{dense}}_{\theta}$ (full attention) and \textbf{sparse policy} $\pi^{\text{sparse}}_{\theta,\phi}$ (sparse attention + LoRA)
\FOR{each RL iteration}
    \STATE Sample rollouts $y \sim \pi^{\text{sparse}}_{\theta,\phi}(\cdot\mid x)$ and compute rewards / advantages
    \STATE Recompute token log-probs with the dense model (FSDP): $\log \pi^{\text{dense}}_{\theta}(y_t\mid x,y_{<t})$
    \STATE Compute token log-probs with the sparse model: $\log \pi^{\text{sparse}}_{\theta,\phi}(y_t\mid x,y_{<t})$
    \STATE Update the base model $\theta \leftarrow \theta - \eta\,\nabla_{\theta}\,\mathcal{L}_{\text{PPO}}\big(\theta; x,y, \log \pi^{\text{dense}}_{\theta}\big)$
    \STATE Freeze $\theta$ and update LoRA $\phi \leftarrow \phi - \eta_{\phi}\,\nabla_{\phi}\,\mathcal{L}_{\text{LoRA}}\big(\phi; x,y, \theta\big)$
\STATE Use LoRA loss $\mathcal{L}_{\text{LoRA}} = \mathcal{L}_{\text{PPO\_Sparse}} + \lambda\,\mathcal{L}_{\text{KL}}$
    \STATE Refresh inference engine weights with updated base model $\theta$ and LoRA $\phi$
\ENDFOR
\end{algorithmic}
\end{algorithm} 

\subsubsection{How \distillsparse Breaks the Vanilla Sparse Rollout Efficiency Threshold} 

For our proof-of-concept case study, we run the experiment with the Qwen3-1.7B Thinking model. 
With the goal of achieving training stability, we start with the 0.86 dynamic sparsity schedule but with LoRA activated. 
For every iteration trained, both the dense policy model and the sparse LoRA are updated, which further boosts the distribution alignment of the 0.86 sparsity (speedup 2.2x) with dense attention. 

Shown in Figure \ref{fig:original_vs_distillsparse_step90}, after training for 20k training examples (phase 1), the green solid and dashed curves show the effect of added LoRA weights and on-policy distillation on the sparse rollout under the original 0.86 sparsity setting. 
We see that through distillation, the same sparsity level achieves significantly higher distribution alignment across all generation sequence lengths. (green dashed is higher than green solid) 
More importantly, we show that \distillsparse generally boosts the 5-percentile acceptance rate across other sparse-attention settings: the original 0.83 setting (coral, speedup 2.3x) and 0.80 setting (blue, speedup 2.5x) also achieve higher distribution alignment after distillation. 
Surprisingly, the original sparsity level of 0.80 now achieves a tail acceptance rate around the 0.86 threshold, and we switch to sparsity of 0.80 from that point forward. 
We show in Figure \ref{mut} (c) that with this switch, the training is still stable and on par with dense for an extended number of steps. More importantly, for steps after 20K training examples, we can train for 2.5x speedup in the rollout, higher than in the first 20K training examples. 
We also show full empirical validation that \distillsparse matches dense rollout in Table \ref{tab:bigbigeval}. 

\textbf{Low Overhead} - Moreover, \distillsparse can be readily integrated into our existing dynamic sparsity scheduling implementation with minimal overhead on top of the original sparse rollout speedup. Details are presented in Appendix \ref{empiricallora}. 

\section{Conclusion} 

Sparse attention can substantially reduce rollout cost in long-chain-of-thought RL, but naive sparse rollout introduces actor--policy mismatch severe enough to destabilize training. In this work, we show that this instability is governed less by average distribution shift than by the tail behavior of a small subset of misaligned tokens, which motivates using lower-percentile acceptance-rate statistics to characterize the true training-relevant divergence. Based on this view, we derive practical sparsity schedules that preserve stable dense-policy training across model scales while still delivering meaningful rollout speedups. We further show that \distillsparse\ improves sparse--dense alignment with minimal overhead, enabling even more aggressive sparse rollout settings to remain trainable. Together, these results provide a practical recipe for making sparse rollout both efficient and stable in RL for large language models.

\bibliographystyle{assets/plainnat}
\bibliography{references} 

\newpage 
\appendix

\section*{Appendix Contents} \phantomsection \label{app:contents} \noindent This appendix provides additional analyses, implementation details, and supporting empirical evidence for the main paper. The sections are organized as follows: \begin{description}[leftmargin=2.2em, style=nextline] \item[{\makebox[\linewidth][l]{\hyperref[spdnhighreward]{Appendix A: Sparse Rollout Instability and High-Reward Rollouts}\hfill p.~\pageref{spdnhighreward}}}] We examine whether insufficient rollout quality is the primary cause of sparse rollout RL instability, and show that this factor alone does not explain training collapse. \item[{\makebox[\linewidth][l]{\hyperref[extended_related_works]{Appendix B: Extended Related Work}\hfill p.~\pageref{extended_related_works}}}] We provide a more detailed discussion of prior work on RLVR, sparse attention, efficient long-context inference, and policy mismatch in reinforcement learning. \item[{\makebox[\linewidth][l]{\hyperref[costbreakdown]{Appendix C: Cost Model Implementation and Visualization}\hfill p.~\pageref{costbreakdown}}}] We describe the implementation details of our rollout cost model and provide additional visualizations illustrating the efficiency tradeoffs of sparse attention. \item[{\makebox[\linewidth][l]{\hyperref[empiricallora]{Appendix D: Empirical Overhead of \distillsparse LoRA Distillation}\hfill p.~\pageref{empiricallora}}}] We quantify the additional training overhead introduced by LoRA-based distillation and analyze its practical impact on end-to-end efficiency. \item[{\makebox[\linewidth][l]{\hyperref[kldivergencedemo]{Appendix E: Distribution Mismatch under Sparse Rollout Collapse}\hfill p.~\pageref{kldivergencedemo}}}] We present additional evidence characterizing the distributional mismatch between sparse and dense rollout policies during unstable RL training. \item[{\makebox[\linewidth][l]{\hyperref[app:rollout_cost_sparse]{Appendix F: Rollout Cost Analysis and Sparse Attention Motivation}\hfill p.~\pageref{app:rollout_cost_sparse}}}] We further motivate sparse attention rollouts through a detailed analysis of rollout-stage computational cost. \end{description}

\section*{Limitations}
\phantomsection
\label{app:limitations}
Our study focuses on sparse rollout for RL with verifiable rewards on a limited set of base models, tasks, and hardware settings, so the exact stability thresholds and speedup tradeoffs we report may not transfer unchanged to other model families, reward formulations, or deployment environments. In particular, our conclusions are tied to long-context reasoning workloads and to the sparse-attention configurations we evaluate, and broader validation across additional domains such as multimodal RL, alternative policy optimization algorithms, and more diverse system stacks would be needed before treating the proposed acceptance-rate thresholds and scheduling rules as universal design principles.

\section*{Broader Social Impact}
\phantomsection
\label{app:broader-social-impact}
If successful, our method can reduce the compute cost of RL training for large language models, which may lower the barrier to studying reasoning systems and make experimentation more accessible to smaller research groups; at the same time, cheaper post-training can also accelerate the development and deployment of increasingly capable models, including systems that may be misused for generating persuasive misinformation, scalable spam, or unsafe automated assistance. We therefore view this work primarily as a systems contribution whose social impact depends on how downstream models are evaluated and released, and we encourage practitioners to pair efficiency gains with careful capability assessment, application-specific safeguards, and responsible release practices. 

\section{Insufficient Rollout Is Not the Main Culprit of Sparse Rollout RL Instability} 
\label{spdnhighreward} 

\begin{figure}[ht]
  \centering
  \includegraphics[width=0.8\textwidth]{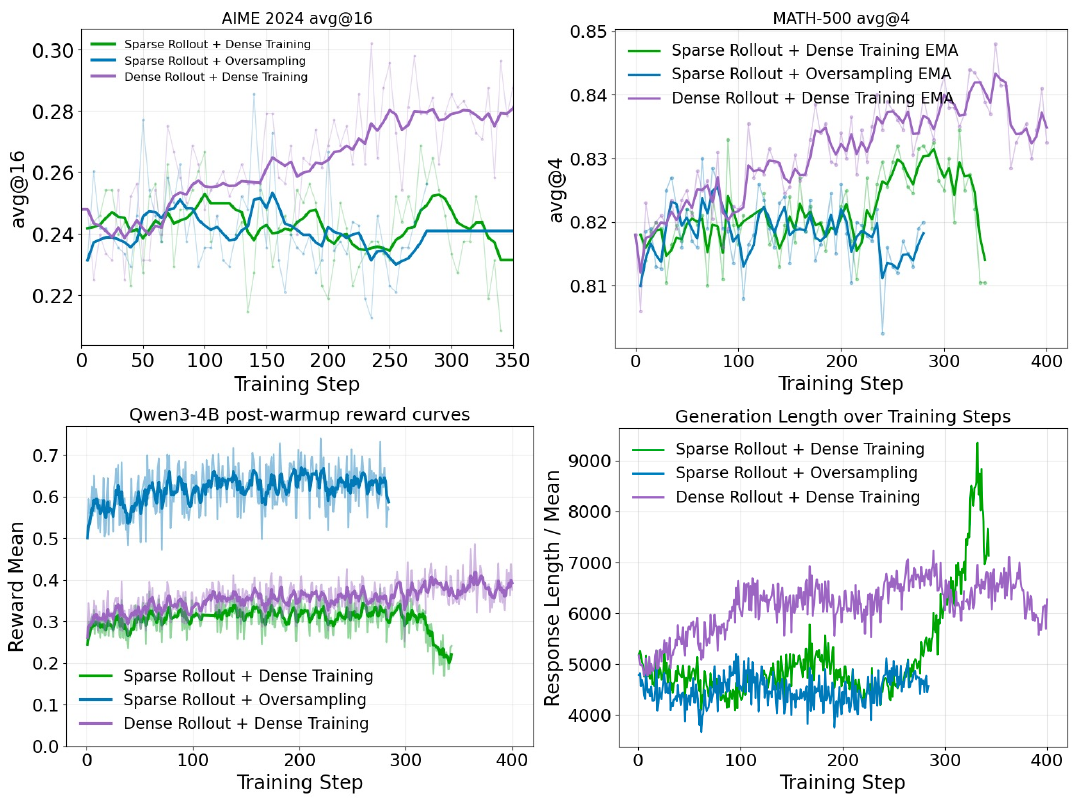} 
  \caption{We illustrate with a simple experiment that sparse rollout RL training dense policy collapse is not only due to the reward is insufficient. Oversampling (Sample 16 select Top-8 based on zero-one-reward) drastically increases reward of the rollout but still cannot prevent the training performance to be near the dense rollout.} 
  \label{fig:superhighr} 
  
\end{figure} 
In this section, we present a concrete study showcasing that although sparse rollout decreases the average RL training reward, the insufficient reward alone does not cause the RL training to crash. 

To save computational resources, we do not use the thinking models as we did in the main experiment section, as training on 37K generation length makes rollout extremely cost-heavy. Instead, we base our experiments on top of base models. To make the model CoT generation length longer, we first warm up the standard base models on the DeepScaleR dataset~\cite{deepscaler2025} for 20K training examples so that the model average generation length rises to above 6K tokens, from where we then train the model on POLARIS dataset~\cite{Polaris2025}. 

Specifically, for the Qwen3-4B-Base model, we perform the above warmup and then start the training with sparse rollout with 12K generation length cutoff. 
We present RL training trials in Figure~\ref{fig:superhighr}. We use \texttt{pagesize} 16 and \texttt{number of pages} 32 for sparse rollout. We compare sparse-rollout RL training (green) with the dense-rollout baseline (purple). On AIME2024 and MATH500, the green curve is consistently below the purple curve. Also note that the sparse-rollout training reward is initially only marginally lower than dense, but eventually crashes. 

For sparse rollout, we attempted to increase its average training rewards. In fact, for every training example problem we sample 16 times (instead of N = eight times in the green curve), then we rank the 16 rollout trajectories per example problem by their training reward and select the top 8. The average reward curve is shown in blue, which is significantly higher than dense average reward and that of the original sparse rollout. However, the blue curve still fails to recover the training performance. 

\section{Extended Related Works} 
\label{extended_related_works} 
We would like to divide the discussion of the related work into four aspects: RL for LLMs, prior works on distribution alignment in RL, general rollout speedup methods, and sparse attention applications in LLMs. 

\textbf{RL for LLMs.} Reinforcement learning has been widely applied to LLMs to improve human alignment, reasoning, coding, and other complex tasks. Beyond PPO, memory-efficient methods have been proposed, including ReMax~\citep{li2023remax}, RLOO~\citep{ahmadian2024back}, and GRPO~\citep{shao2024deepseekmath}. In addition, methods such as SimPO~\citep{meng2024simpo} and DPO~\citep{rafailov2023direct}, which are based on offline RL, have also been employed for human alignment. RL training systems for LLMs, such as Verl~\citep{sheng2025hybridflow}, AReal~\citep{fu2025areal}, TRL~\citep{vonwerra2022trl}, and OpenRLHF~\citep{hu2024openrlhf}, have been developed to improve training throughput and scalability.

\textbf{Distribution Mismatch Correction in RL.} Actor-policy mismatch is a common problem that has long been studied, e.g. Impala \citep{espeholt2018impalascalabledistributeddeeprl}. To alleviate the actor-policy distribution gap, the method introduces a truncated importance sampling (TIS) to approximate the true PPO objective. 
\begingroup 
\begin{equation}
    \mathcal{L}^{\text{PPO}}(\theta) 
    = \mathbb{E}_{x \sim P_{\text{inf}}}
    \Big[ \textcolor{red}{\min \big(\tfrac{p_{\text{ref}}(x)}{p_{\text{inf}}(x)}, C \big)}\min \big( r_\theta(x)\,\hat{A}(x), 
    \operatorname{clip}(r_\theta(x), 1-\epsilon, 1+\epsilon)\,\hat{A}(x) \big) \Big]
\end{equation} 
\endgroup 

The truncation threshold $C$ maintains the stability of the range of the importance ratio. Recently, several methods apply the truncated importance sampling method to RL of LLMs. 
Methods such as FlashRL~\citep{liu2025flashrl}, AReal~\citep{fu2025areal}, and LlamaRL~\citep{wu2025llamarldistributedasynchronousreinforcement} address distribution mismatch by introducing (truncated) importance sampling ratios, typically of the form $p_{\text{ref}}/p_{\text{inf}}$, to correct the impact of mismatch on advantage estimation. From a system perspective, FP32 LM heads~\citep{liu2025flashrl} and deterministic LLM inference~\citep{he2025nondeterminism} are implemented to mitigate the numerical issue of serving systems during rollout. 

\textbf{Prior Rollout Speedup Methods.} 
Many recent works have been proposed to address this rollout efficiency challenge, but have several key limitations.
Several recent works~\citep{zheng2025prosperity,piché2025pipelinerlfasteronpolicyreinforcement,zhou2025aprilactivepartialrollouts} have designed asynchronous RL training systems that accelerate training by decoupling the rollout and training phases to better utilize computational resources. 
However, although this line of work prioritizes training throughput, the high resource utilization in the rollout phase remains unaddressed.
Moreover, the asynchrony can result in staleness of samples, which can potentially lead to inferior training.
Another line of work aims to accelerate rollouts with model quantization~\citep{liu2025flashrl,huang2025qerlefficiencyquantizationenhanced} and speculative decoding~\citep{leviathan2023fastinferencetransformersspeculative, chen2023acceleratinglargelanguagemodel}.
Although model quantization can significantly reduce the cost of loading model weights, it cannot effectively mitigate the rollout overhead for long-sequence generation, where KV-cache loading remains the primary bottleneck~\citep{sadhukhan2025kineticsrethinkingtesttimescaling}.
Conversely, speculative decoding can accelerate rollouts without altering the sampling distribution. However, it is largely unsuitable for large-batch rollout settings \citep{liu2025turbospecclosedloopspeculationcontrol,su2023synergyspeculativedecodingbatching} in RL training because the verification process becomes compute-intensive. Furthermore, speculative decoding introduces an additional draft model that requires extra training resources and thus complicates the whole training pipeline. 

\textbf{Sparse attention.} Attention-operation cost dominates the latency of generating long-context output, a consensus shared by many prior studies \citep{sadhukhan2025kineticsrethinkingtesttimescaling,yuan2025nativesparseattentionhardwarealigned}. A substantial body of work has focused on reducing the attention cost by pruning redundant token computation and loading and computing only essential tokens. Training-free approaches revolve around fine-grained token-level decisions \citep{zhang2023h2oheavyhitteroracleefficient} or more accurate dynamic block-sparse attention \citep{tang2024questqueryawaresparsityefficient,sun2024triforcelosslessaccelerationlong,liu2025clusterkvmanipulatingllmkv}.
Despite robust performance in general tasks, under aggressive sparsity settings, these methods incur an unacceptable accuracy drop. Pretrained sparse attention methods \citep{yuan2025nativesparseattentionhardwarealigned,deepseekai2024deepseekv32}, on the other hand, achieve scalable results. 

\section{Detailed Implementation and Visualization of the Cost Model} 
\label{costbreakdown} 
\begin{figure}[ht]
  \centering
  \includegraphics[width=0.7\textwidth]{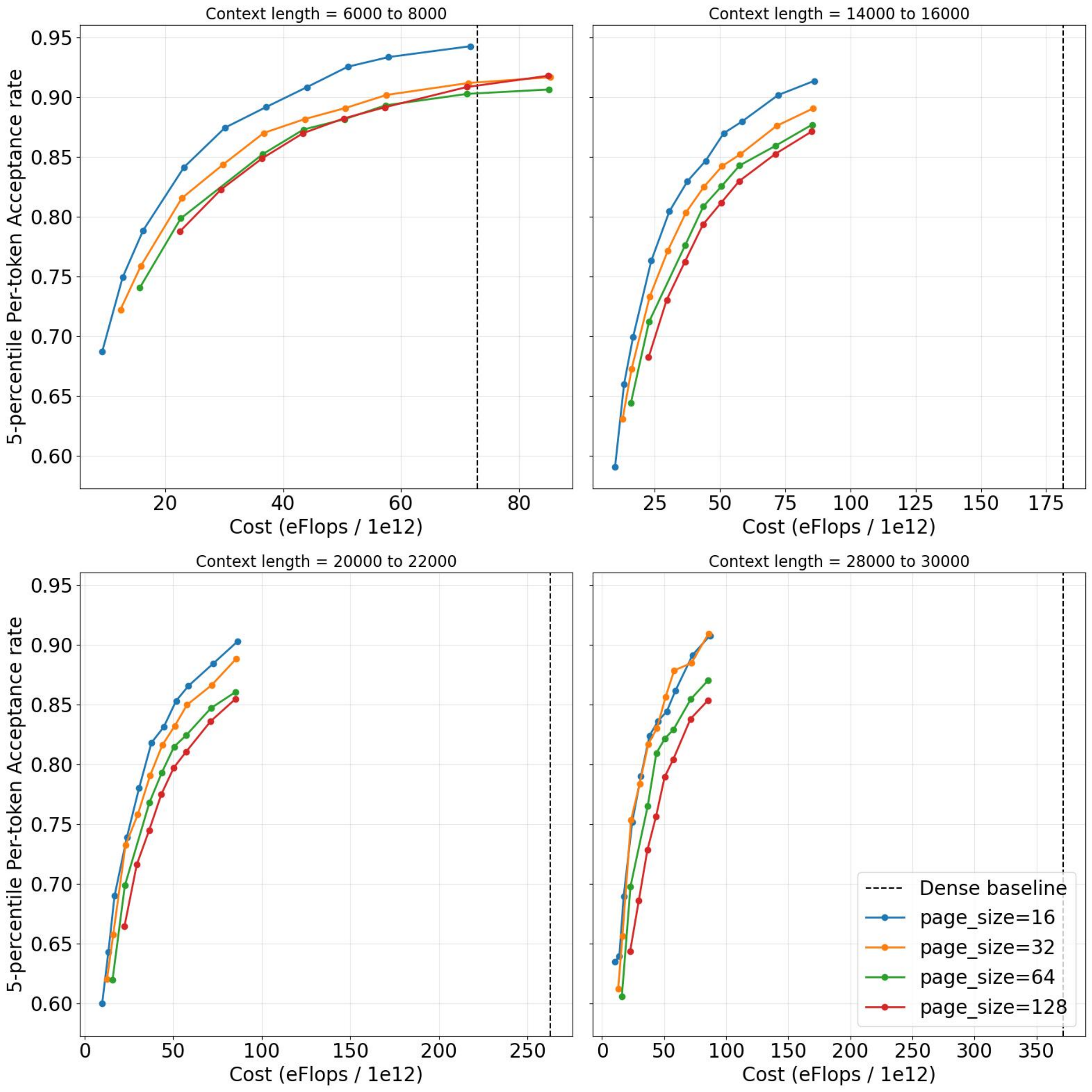} 
  \caption{Additional cost analysis showing how different page sizes compare across generation-length regimes.} 
  \label{fig:costvisualization} 
  
\end{figure} 

\begin{wraptable}{r}{0.42\textwidth}
\centering
\footnotesize 
\caption{Empirically, we also observe a similar pattern: the page size from 16 and above does not affect the decoding speed much as long as the sparse engine's Top-$k$ kernel is well implemented (Cost measured on single H200 with batch size 16 in ms).} 
\label{tablecostpage} 
\begin{tabular}{c|ccc}
\toprule 
Vortex configs & \multicolumn{3}{c}{Past KV Size} \\
(\texttt{pagesize}-\texttt{num\_page}) & 16k & 24k & 32k \\
\midrule
166-4 & 4.59 & 5.13 & 5.41 \\
32-32 & 4.47 & 4.96 & 5.30 \\
64-16 & 4.30 & 4.87 & 5.12 \\
128-8 & 4.13 & 4.84 & 5.07 \\
\bottomrule
\end{tabular}
\end{wraptable} 

In this section, we provide more details on the cost-model implementation and visualize how \texttt{pagesize} relates to cost and distribution mismatch with the dense model. 

Specifically, for the cost model analysis, we use Qwen3-1.7B model configurations for now. 
Generalizations to bigger models will be presented in later iterations. 
In particular, we set \texttt{parallel\_sampling} to 8, mimicking the conventional GRPO rollout scenario, and base the analysis on one H200, similar takeaways can also be drawn on B200. (Specifically, we use arithmetic intensity based on H200 but for our conclusion the arithmetic intensity holds when we use B200 arithmetic intensity.) 
For each \texttt{pagesize} setting, we pick 11 KV-budget configurations, covering 0.7 to 0.9 tail mismatch across all experimental sequence lengths. 

There are two aspects that we think are worth highlighting. 
First, \texttt{pagesize}=16 occupies the Pareto frontier across all generation lengths. 
However, as generation length increases, the gap between \texttt{pagesize}=32 and 16 shrinks; by 30K tokens, \texttt{pagesize}=32 nearly reaches the Pareto frontier and is almost on par with \texttt{pagesize}=16 in the cost-mismatch tradeoff. 
Second, as the generation sequence length increases, the gap between the cost of using sparse attention rollout and dense attention increases, showing more cost benefits from using sparse attention. 

Benefiting from the high-quality top-$k$ kernels supported in \cite{vortex_torch2026}, we are able to empirically show that the scoring portion of the sparse attention cost in the model generation is not significantly impacted by the page size selection. 
Shown in Table~\ref{tablecostpage}, we measure the average generation cost in the full forward pass of the sparse attention model under different \texttt{pagesize} selections. 
Particularly, we use a single H200 GPU, and we prefill with sequence lengths 16K, 24K, and 32K with batch size 8, and compute the average full-model decoding pass by averaging over 1000 trials. 
The average is presented in milliseconds (ms). 

\section{Empirical Study of \distillsparse Overhead of LoRA Distillation Training} 
\label{empiricallora} 

\begin{figure}[ht]
  \centering
  \includegraphics[width=0.6\textwidth]{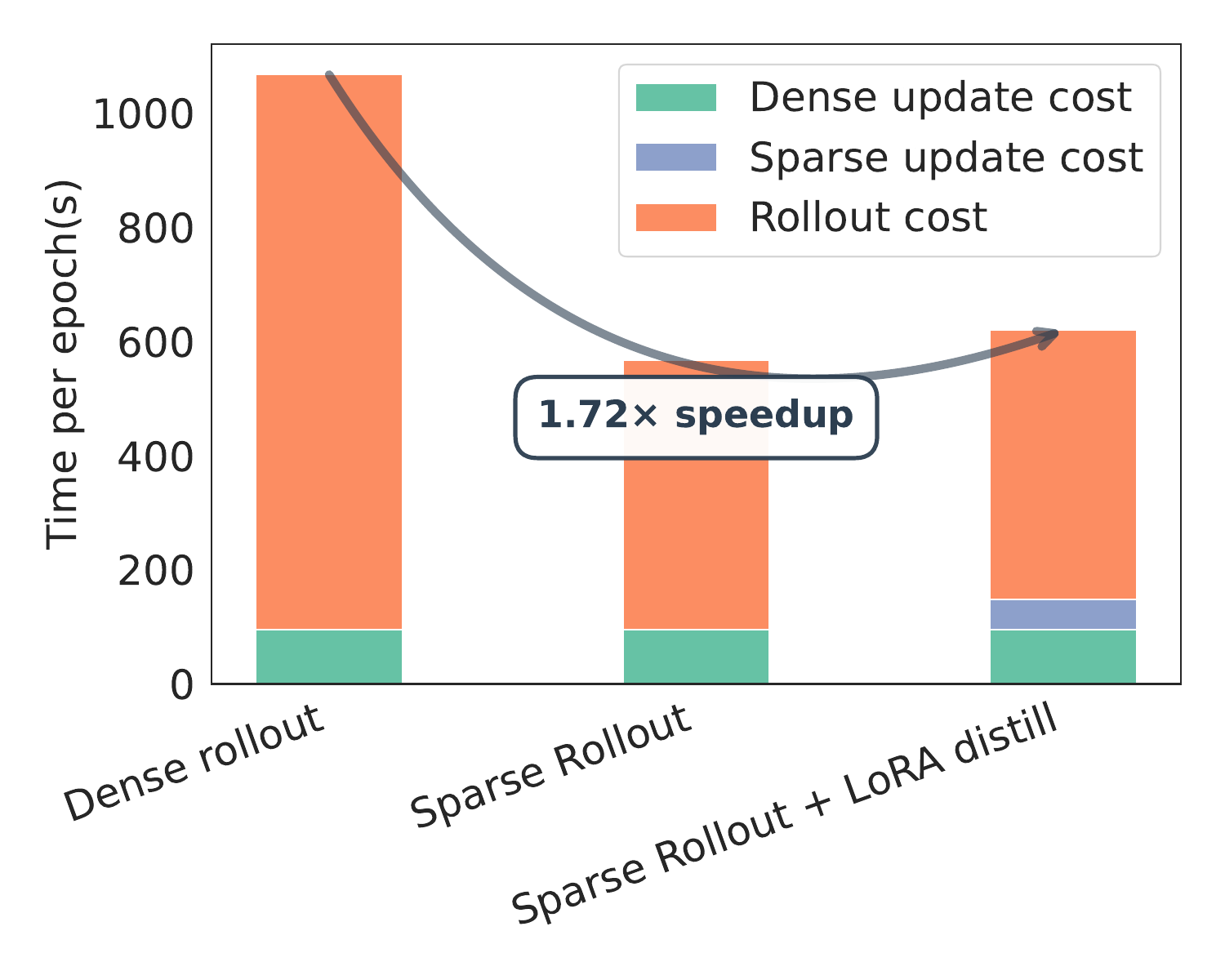} 
  \caption{}
  \label{fig:speedup}
  
\end{figure} 

\textbf{System implementation \& Efficiency:} 
In this section, we show that \distillsparse incurs minimal overhead. 
To simplify the implementation, for the speedup demonstration, we choose a simplified setting, where instead of using a sparsity schedule for rollout and training with dense distillation, we use a fixed sparse budget. 

To conduct our experiments, we use Verl~\citep{sheng2024hybridflow} with FSDP as the training engine and SGLang~\citep{zheng2024sglangefficientexecutionstructured} as the inference engine. Experiments are run on Qwen3-4B-Instruct with generation length 16K. 
Training is run on 2xH200 GPUs. 
For efficient sparse-attention rollouts, we use Vortex\_torch~\citep{vortex_torch_docs}. We adopt block top-$k$ attention with a page size of 16, and set the number of top-$k$ pages according to the sparse KV budget. In addition, we use Flash Sparse Attention~\citep{yan2025flashsparseattentionalternative} for efficient sparse-attention training and PEFT~\citep{peft} for LoRA adaptation. 

As shown in Figure \ref{fig:speedup}, we report the efficiency of our implementation. When training a 4B instruct model with 16K max context length, dense rollouts account for roughly 90\% of the per-epoch time. Sparse attention directly alleviates this bottleneck and accelerates rollouts by roughly $1.9\times$. Although the dense policy update contributes only about 10\% of the total cost, our sparse distillation increases this component by approximately 55\%. Overall, we obtain a $1.72\times$ end-to-end speedup, showing the promising effect of small overhead. 

\section{Distribution Mismatch under Sparse Rollout Training Collapse: Visualization of KL Divergence of Actor-Policy Mismatch} 
\label{visualizationmismatch} 

\begin{figure}[!ht]
  \centering
  \includegraphics[width=0.6\textwidth]{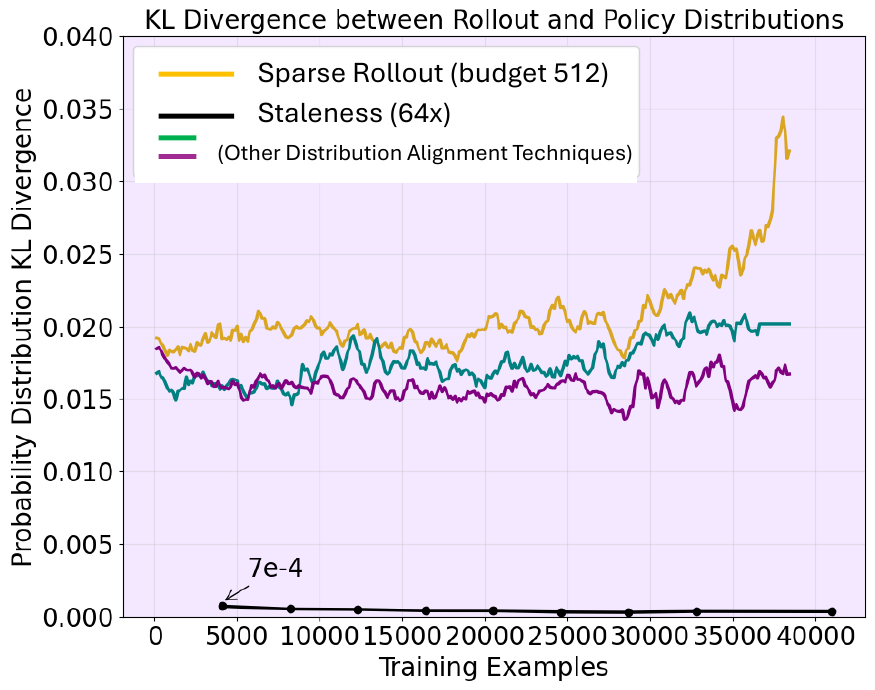}
  \caption{Comparison of actor-policy KL divergence for Qwen3-4B-Base models under 64 times staleness and sparse-to-dense training.} 
  \label{kldivergencedemo} 
  
\end{figure} 

In this section, we show an illustration of why sparse-dense RL training causes such a huge problem when selecting sparsity in an ad hoc manner. Compared to traditional scenarios such as aggressive staleness where distribution alignment techniques like TIS~\citep{liu2025flashrl} and \citep{jackpot2025github} are used, choosing aggressive sparsity for rollout makes the actor-policy divergence much more severe; in particular, the KL divergence is orders of magnitude higher. 

Specifically, we measure the above visualization under the following settings. To save compute, we follow Section~\ref{spdnhighreward} and use Qwen3-4B-Base after warmup with sparse rollout (KV budget 512). 
With TIS + Jackpot applied, the resulting KL divergence is plotted in the yellow curve of Figure~\ref{kldivergencedemo}, where it crashes in the later stage of training. 
In contrast, we also plot the KL divergence of the 64x staleness on Qwen3-4B-Base in black. 

\begin{figure}[!ht]
  \centering
  \includegraphics[width=0.6\textwidth]{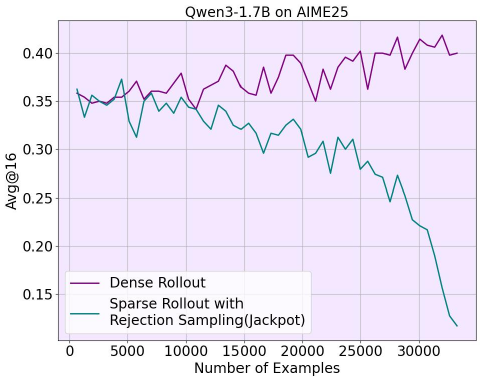} 
  \caption{Besides importance-sampling alignment methods, we show that rejection sampling (Jackpot) is also unable to solve training instability of 37K long-context rollout by itself.} 
  \label{jptweaknesses} 
  
\end{figure} 

The huge KL divergence renders the conventional alignment techniques unable to offer a general solution for sparse rollout training, motivating for a comprehensive study of optimal mismatch and optimal sparsity configuration selection. 
In the paper, we show in Figure \ref{fig:tradeoffandmismatchcost}(a) that Qwen3-1.7B thinking under sparse rollout with a KV budget of 1024, TIS alone cannot solve the issue of training instability. 
Here in Figure \ref{jptweaknesses}, we also provide the study of using the rejection-sampling method \cite{jackpot2025github} in the same setting, and show that it also cannot repair instability.


\section{Rollout Cost Analysis and Sparse Attention Motivation}
\label{app:rollout_cost_sparse} 

To understand why rollout generation dominates RL training, prior work models decoding cost as the combination of computation and KV-cache memory traffic~\citep{sadhukhan2025kineticsrethinkingtesttimescaling}. We adopt the same additive view and analyze rollout latency independently.

For dense autoregressive decoding, the per-layer cost consists of parametric computation plus self-attention:
\[
C_{\text{comp}} = 2PBNL_{\text{out}}
\;+\;
rBN(2L_{\text{in}}+L_{\text{out}})L_{\text{out}}D,
\]
while memory traffic is driven by parameter and KV-cache access:
$$
C_{\text{mem}} = 2PL_{\text{out}}
\;+\;
BN(2L_{\text{in}}L_{\text{out}}+L_{\text{out}}^2)D.
$$

Although parameter access can be amortized with large batch size and model parallel inference, the attention term grows quadratically with response length. As a result, rollout decoding becomes increasingly memory-bound for long generations, and self-attention quickly emerges as the dominant bottleneck.

\paragraph{Sparse attention.}
With a KV budget $K$, sparse decoding replaces the quadratic dependence on $L_{\text{out}}$ with a budgeted cost:
\[
C_{\text{comp}}^{(\text{sparse }K)} =
2PBNL_{\text{out}}
+
rBND\cdot H(L_{\text{in}},L_{\text{out}},K),
\]
\[
C_{\text{mem}}^{(\text{sparse }K)} =
2PL_{\text{out}}
+
BND\cdot H(L_{\text{in}},L_{\text{out}},K),
\]
where
\begin{align*}
H&=(2L_\text{in}+L_{\text{dense}})L_{\text{dense}}
+2K\bigl(L_{\text{out}}-L_{\text{dense}}\bigr)_+,
\\
L_{\text{dense}}&=\min(L_{\text{out}},K-L_{\text{in}}).    
\end{align*}

Thus sparse attention avoids $\mathcal{O}(L_{\text{out}}^2)$ scaling and instead yields $\mathcal{O}(K L_{\text{out}})$ memory growth.

\textbf{End-to-End Implications}: An RL iteration decomposes into rollout generation, logits recomputation, and policy update:
\[
C_{\text{total}} = C^{\text{rollout}} + C^{\text{recomp}} + C^{\text{update}}.
\]
Recomputation and update operate on full sequences via prefill-style passes, achieving high arithmetic intensity and remaining largely compute-bound. In contrast, rollout decoding performs per-token KV access with low arithmetic intensity, making it fundamentally memory-bound.

\paragraph{Superior Inference Scaling with Sparse Attention.} Overall, the rollout phase dominates end-to-end training time (often $>90\%$, Figure~\ref{fig:speedup}), primarily because the KV loading cost increases quadratically with generation length~\citep{sadhukhan2025kineticsrethinkingtesttimescaling}. Increasing the number of rollouts cannot amortize the decoding cost as the dominant KV memory also grows linearly with it.
Sparse attention directly targets this bottleneck by reducing rollout memory cost from $\mathcal{O}(L_{\text{out}}^2)$ to $\mathcal{O}(K L_{\text{out}})$, where $K$ is the sparse KV cache size. This allows us to increase the \emph{generation length} and \emph{number of samples} to achieve high-quality rollouts within reasonable cost. In other words, \textbf{sparse attention unlocks superior inference scaling in the usual long CoT regime.} 



\end{document}